\documentclass{article}

\usepackage[preprint]{corl_2026}
\usepackage{graphicx}
\usepackage{amsmath}
\usepackage{amssymb}
\usepackage{booktabs}
\usepackage{capt-of}
\usepackage{multirow}

\title{How to Learn from What a Human Would Avoid? Intervention-Aware World Models with Real-World RL for Dexterous Manipulation}

\author{
  \bf Jiaju Yin$^{1,*}$ \quad Zhenhui Zhang$^{1,*}$ \quad Lixin Xu$^{1}$ \quad Heng Zhang$^{2}$ \\
  \bf Jun Shao$^{3}$ \quad Yating Feng$^{1}$ \quad Arash Ajoudani$^{2,\dagger}$ \quad Renjing Xu$^{1,\dagger}$ \\[3pt]
  \rm $^{1}$HKUST (Guangzhou) \quad
  $^{2}$Italian Institute of Technology \quad
  $^{3}$Zhejiang University \\[3pt]
  \rm $^{*}$Equal contribution \quad $^{\dagger}$Corresponding author
}

\begin{document}
\hypersetup{
  pdftitle={How to Learn from What a Human Would Avoid? Intervention-Aware World Models with Real-World RL for Dexterous Manipulation},
  pdfauthor={Jiaju Yin, Zhenhui Zhang, Lixin Xu, Heng Zhang, Jun Shao, Yating Feng, Arash Ajoudani, Renjing Xu}
}
\maketitle
\vspace{-2.0em}

\vspace{-0.9em}
\begin{center}
\small Project page: \url{https://whirl-dexterous.github.io/}
\end{center}
\vspace{-0.2em}

\vspace{-0.5em}
\noindent
\begin{minipage}{\linewidth}
\centering
\includegraphics[width=1.00\linewidth]{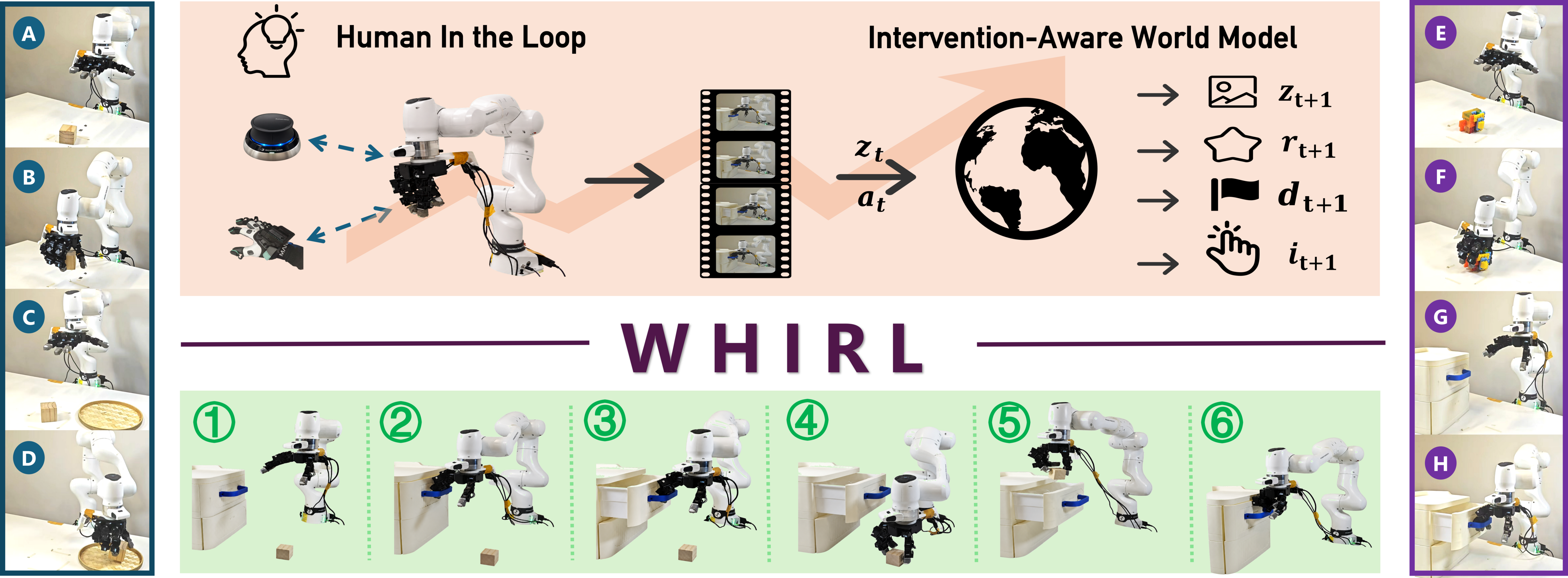}
{\small\captionof{figure}{\textbf{WHIRL learns reusable risk from human takeovers.} Each pedal intervention becomes a next-step takeover label for an intervention-aware world model, whose prediction head shapes the residual actor away from intervention-prone states across dexterous manipulation tasks.}\label{fig:teaser}}
\end{minipage}
\vspace{-1.0em}

\begin{abstract}
Multi-fingered dexterous manipulation remains a frontier for real-world reinforcement learning (RL) due to the high-dimensional action space and the prohibitive cost of hardware failures. While human-in-the-loop (HIL) RL allows operators to intervene before failures occur, current pipelines often treat these interventions as reactive corrections, discarding the rich safety signal inherent in the operator's decision to take control. In this paper, we ask: \emph{How can we learn from what a human would avoid?}

We present \textbf{WHIRL}, a safety-aware RL framework that transforms binary human interventions into forward-predictive signals for proactive risk avoidance. Our approach centers on an intervention-aware latent world model with four prediction heads: dynamics, reward, termination, and a novel per-state intervention-probability head that learns to predict the likelihood of a human takeover at future states. This head provides an actor-side risk-shaping term that discourages the policy from entering ``intervention-prone'' regions, modeling the operator's internal safety threshold.

We evaluate our framework on a 16-DoF LEAP Hand across tasks spanning convex and irregular object grasping, prismatic manipulation, and long-horizon multi-stage tasks. Our results show that predictive risk-shaping enables the system to achieve a 96.7\% success rate on complex grasping tasks while reducing the operator intervention burden by up to 84\% in step-weighted terms. By closing the loop between human intuition and predictive world modeling, this work provides a practical safety-aware recipe for training complex dexterous agents in the real world while reducing operator fatigue and hardware-risk exposure.
\end{abstract}

\vspace{-1.2em}

\clearpage
\vspace{-2.5em}
\section{Introduction}
\label{sec:intro}

Dexterous manipulation~\cite{openai2020dexterous, handa2022dextreme, chen2023visualdexterity}, grasping, repositioning, and placing objects with multi-fingered hands, remains a central challenge in real-world robot learning.
Behavior cloning from teleoperated demonstrations~\cite{zhao2023act, chi2023diffusion} has become the practical starting point: a compact demonstration set can teach a dexterous hand the nominal motion needed to approach, grasp, and place objects.
However, contact-rich manipulation is governed by small variations in object pose, finger timing, and incipient slip.
These variations are difficult to cover exhaustively with demonstrations, so a behavior-cloning policy that appears competent on nominal trials can still fail under moderate deployment shift.

Online reinforcement learning offers a direct way to turn these failures into policy improvement~\cite{rajeswaran2018dapg, johannink2019residual, ball2023rlpd, zhang2024srl}.
For low-dimensional grippers, SERL~\cite{luo2024serl} and related actor--learner systems~\cite{ball2023rlpd, hu2024ibrl} show that online RL can reach strong real-robot performance within hours.
For a 16-DoF dexterous hand, the same recipe becomes substantially more expensive.
The policy acts in a coupled arm--hand action space, contacts are stiff and discontinuous, and each exploratory mistake consumes real robot time, resets, and hardware attention.
Sim-to-real approaches~\cite{openai2019rubiks, qin2023dexpoint, handa2022dextreme} reduce this burden when accurate simulation is available, but they do not remove the need for real-robot adaptation on a new hand, task, and teleoperation stack.

Human-in-the-loop (HIL) RL is a natural compromise: let the policy explore, but allow a human to take over before unrecoverable failures. HIL-SERL~\cite{luo2024hilserl}, SiLRI~\cite{silri2025}, and interactive imitation methods~\cite{kelly2019hgdagger, hoque2022thriftydagger, liu2023sirius} use interventions to stabilize real-robot learning. The cost of HIL is expert attention: each takeover consumes operator time, so the central question is sharp---\emph{does each second of operator attention pay off after the pedal is released?}

Current HIL pipelines \emph{use each takeover once and throw the rest of the signal away}: the binary label becomes a behavior-cloning mask, a replay priority, or a safety metric, but never supervision for predicting where the autonomous policy will soon need help. This points to a simple use of latent world models~\cite{hafner2020dreamer, hafner2025dreamerv3, janner2019mbpo, hansen2022tdmpc}: alongside dynamics, reward, and termination, learn a takeover-probability head that converts each intervention into a reusable actor-side risk signal.

We present \textbf{WHIRL}~(\textbf{W}orld models, \textbf{H}uman-in-the-loop, \textbf{I}ntervention, \textbf{R}eal-world, reinforcement \textbf{L}earning), and make three contributions:

\noindent\textbf{Our contributions.}
\vspace{-0.5em}
\begin{itemize}
\setlength\itemsep{1pt}\setlength\parskip{0pt}\setlength\topsep{2pt}
\item \textbf{Intervention-aware world model.} A compact latent model predicts dynamics, reward, termination, and next-step takeover probability; the first three heads provide uncertainty-gated critic targets, while the intervention head supplies actor-side risk shaping.
\item \textbf{Mid-rollout HIL teleoperation system.} A tri-channel interface couples arm control, glove-based hand control, and pedal takeover; its LEAP\,+\,Manus retargeter combines non-thumb affine joint mapping, thumb fingertip IK, and command rebase for continuous interventions.
\item \textbf{Controlled real-robot evidence.} On a 16-DoF LEAP Hand, we evaluate WHIRL across five dexterous tasks with a progressive baseline ladder and a mechanism ablation, isolating how predictive intervention modeling reduces operator burden.
\end{itemize}

\begin{figure}[t]
\centering
\includegraphics[width=\linewidth]{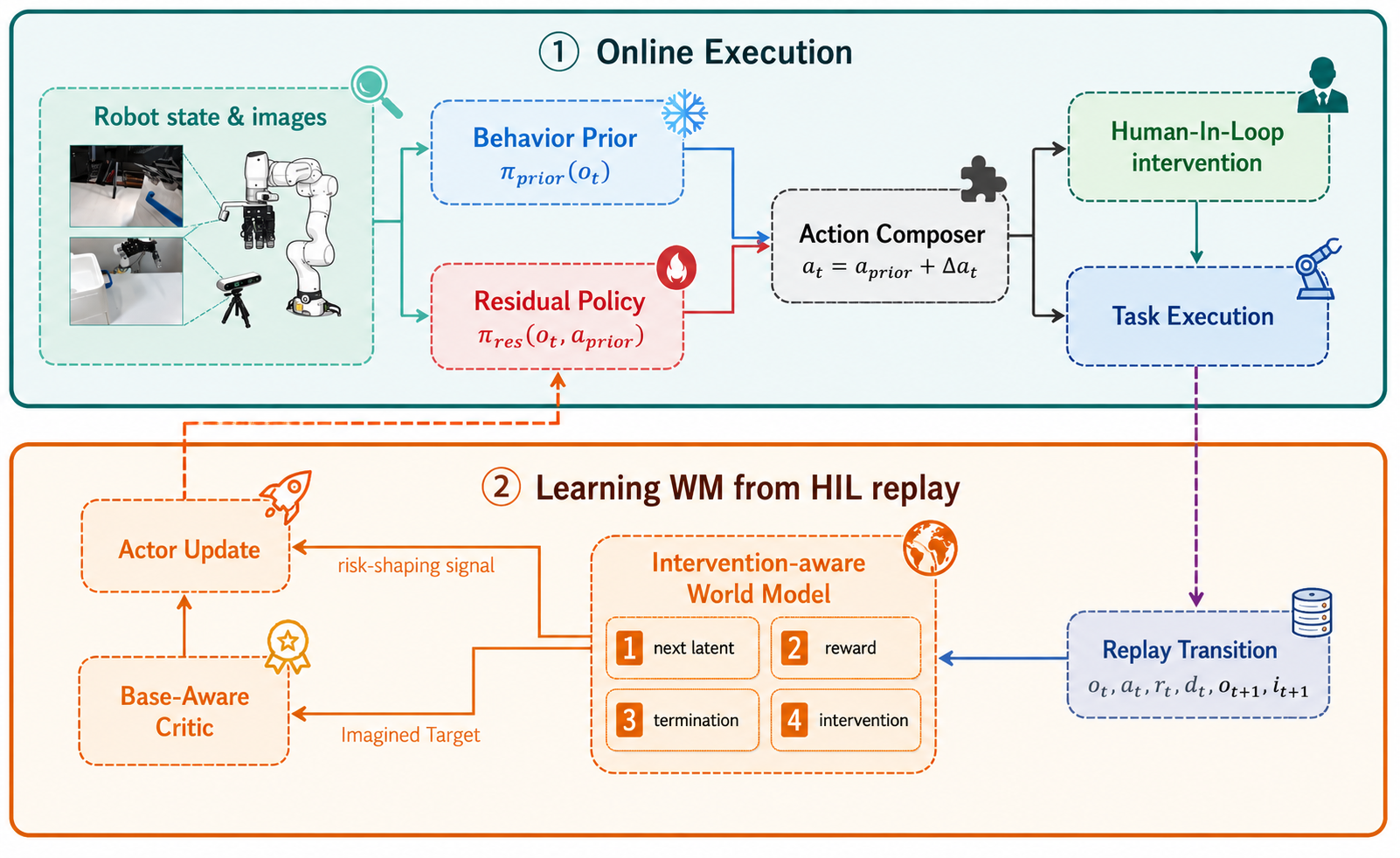}
{\small\caption{WHIRL overview: \textbf{takeovers become predictive risk.} During online execution, a frozen behavior prior and residual policy compose arm--hand actions while HIL takeovers create labeled replay. The intervention-aware world model then predicts dynamics, reward, termination, and next-step intervention probability; the first three heads support critic learning, while the intervention head supplies actor-side risk shaping.}\label{fig:system}}
\end{figure}

\section{Related Work}
\label{sec:related}

\paragraph{Real-world online RL for manipulation.}
Real-world online RL has advanced through direct actor--learner systems such as SERL~\cite{luo2024serl} and related platforms~\cite{ball2023rlpd, hu2024ibrl}, and more recently through large-policy post-training such as RECAP ($\pi^{*}_{0.6}$)~\cite{pi06} and Fleet-Scale RL~\cite{wang2026fleetrl}. For dexterous hands, residual RL is a conservative alternative: ResFiT~\cite{ankile2025resfit} freezes a behavior-cloning prior and learns a sparse-reward correction, providing the backbone we adopt. HIL methods, such as HIL-SERL~\cite{luo2024hilserl}, SiLRI~\cite{silri2025}, and interactive imitation~\cite{kelly2019hgdagger, hoque2022thriftydagger, liu2023sirius}, make exploration safer by adding takeovers to replay, but use each takeover once and throw the rest of the signal away: the label becomes current-step behaviour supervision, a replay weight, or a metric. WHIRL instead trains it as a next-step intervention-risk prediction target the actor consumes on every state.
\vspace{-0.5em}
\paragraph{World models for robot control.}
Dreamer~\cite{hafner2020dreamer, hafner2025dreamerv3}, MBPO~\cite{janner2019mbpo}, and TD-MPC~\cite{hansen2022tdmpc} established latent world models that predict environment quantities and train actors or critics with imagined data. Hi-WM~\cite{li2026hiwm} moves this idea into HIL learning by letting the operator intervene inside a learned model. WHIRL instead keeps corrections on the real robot and uses a small one-step world model whose distinctive head predicts future human takeover. Dynamics, reward, and termination support the critic, while the intervention head supplies actor-side risk shaping.
\vspace{-0.5em}
\paragraph{Dexterous teleoperation and synergies.}
Dexterous teleoperation systems retarget human hand motion through fingertip objectives, joint-space mappings, or optimization~\cite{qin2023anyteleop, handa2020dexpilot, wang2024dexcap, shaw2024bimanual}; DexRetargeting~\cite{qin2023anyteleop} and Bidex Manus Teleop~\cite{shaw2024bimanual} are the closest general-purpose pipelines for our LEAP\,+\,Manus rig. WHIRL adapts this line of work to mid-rollout HIL with a morphology-routed retargeter, non-thumb affine joint mapping plus thumb fingertip IK, and command rebasing at each takeover. Following hand-synergy priors~\cite{santello1998postural, ciocarlie2009hand, ciocarlie2007eigengrasps}, WHIRL restricts the online residual to a low-dimensional coordination subspace.

\section{Method}
\label{sec:method}

WHIRL combines three components illustrated in Figure~\ref{fig:system}: a tri-channel HIL teleoperation stack that produces clean intervention labels (Section~\ref{sec:method-teleop}), a residual-RL backbone over a behavior prior with hand exploration restricted to a low-dimensional synergy basis (Section~\ref{sec:method-residual}), and a 4-head intervention-aware world model that turns takeover labels into actor-side risk shaping (Section~\ref{sec:method-wm}).

\subsection{Human-in-the-Loop Teleoperation System}
\label{sec:method-teleop}

WHIRL collects HIL data with a tri-channel teleoperation stack that supports continuous mid-rollout takeovers and logs each pedal state as an intervention label.

\textbf{Tri-channel control.} A 3Dconnexion SpaceMouse drives the Franka end-effector, a Manus Quantum glove drives the hand, and a foot pedal toggles between autonomous execution and human teleoperation. The pedal state simultaneously labels every transition with a binary indicator $i_t$ that serves both as an evaluation metric and as supervision for the world-model intervention head.

\textbf{Morphology-routed hand retargeter.} On our LEAP\,+\,Manus rig, per-frame full-hand fingertip IK (e.g.\ DexRetargeting~\cite{qin2023anyteleop}) is robust but introduces visible latency at the 120\,Hz glove rate, while direct joint-space mapping (e.g.\ Bidex Manus Teleop~\cite{shaw2024bimanual}) is zero-latency but loses thumb--index opposition. We therefore route morphology-by-morphology: the non-thumb fingers (index, middle, ring) use a per-joint affine map fit once from three calibration postures, while the 4-DoF thumb is solved by workspace fingertip IK augmented with an \emph{opposition-direction term} that recovers the power-grasp wrap position-only tracking misses.

\textbf{Intervention rebase for continuity.} Let $t_0$ be the takeover time, $\mathbf{g}_t$ the Manus glove state, $R(\cdot)$ the per-frame hand retargeter, and $\mathbf{q}^{L,\mathrm{robot}}_{t_0}$ the LEAP joint state at takeover. We remove command jumps with a $\Delta$-joint rebase:
\begin{equation}
    \mathbf{q}^{L,\mathrm{cmd}}_t = \mathbf{q}^{L,\mathrm{robot}}_{t_0} \;+\; \bigl(R(\mathbf{g}_t) - R(\mathbf{g}_{t_0})\bigr),
    \label{eq:teleop_rebase_main}
\end{equation}
Thus the first command equals the current robot hand pose, and later commands apply only the retargeted glove displacement since takeover. The rebase is independent of the particular retargeter.

\subsection{Residual RL Based on a Behavior Prior}
\label{sec:method-residual}

WHIRL performs residual reinforcement learning~\cite{silver2018residual, johannink2019residual} over a frozen chunk-based behavior prior $\pi_\mathrm{BC}$ fit to teleoperated demonstrations. The executed action is $\mathbf{a}_t = \mathbf{a}^{\mathrm{BC}}(\mathbf{o}_t) + \Delta_\theta(\mathbf{o}_t)$, and only the residual $\Delta_\theta$ is updated online. The replay buffer $\mathcal{D}=\mathcal{D}_{\mathrm{demo}}\cup\mathcal{D}_{\mathrm{auto}}\cup\mathcal{D}_{\mathrm{int}}$ stores autonomous and intervention transitions with pedal labels $i_t\in\{0,1\}$; intervention transitions store the operator command rather than the policy-composed action. Section~\ref{sec:method-wm} turns these labels into takeover-prediction targets, and Section~\ref{sec:exp-ablation} isolates their actor-side effect.

Random perturbations in the raw 16-DoF hand joint space mostly break the coordinated postures the prior has learned. We therefore restrict the online hand residual to a low-dimensional postural synergy basis~\cite{santello1998postural, ciocarlie2007eigengrasps} $\mathbf{W}\!\in\!\mathbb{R}^{K\times 16}$ shared across tasks. The residual is
$\Delta_\theta(\mathbf{o}_t) = \bigl[\Delta\mathbf{a}^{\mathrm{arm}}_\theta(\mathbf{o}_t),\; \mathbf{W}^{\top}\mathbf{z}^{\mathrm{hand}}_\theta(\mathbf{o}_t)\bigr]$, which maps the low-dimensional hand update back to the LEAP joint space. The actor is optimized to improve return while remaining a local correction around the prior:
\begin{equation}
    \mathcal{L}_{\mathrm{actor}} = -\mathbb{E}_{\mathbf{o}_t \sim \mathcal{D}}\!\left[ Q_\phi\!\left(\mathbf{o}_t,\, \mathbf{a}^{\mathrm{BC}}_t + \Delta_\theta(\mathbf{o}_t)\right) \right] + \alpha_{\mathrm{arm}}\|\Delta \mathbf{a}^{\mathrm{arm}}_\theta\|_2^2 + \alpha_{\mathrm{hand}}\|\mathbf{z}^{\mathrm{hand}}_\theta\|_2^2 ,
    \label{eq:actor_loss}
\end{equation}
with the hand regularizer applied in synergy space, so each penalty unit corresponds to a coordinated postural change instead of a single-joint deviation.

\subsection{Intervention-Aware World Model}
\label{sec:method-wm}

\begin{figure}[h!]
\centering
\includegraphics[width=\linewidth]{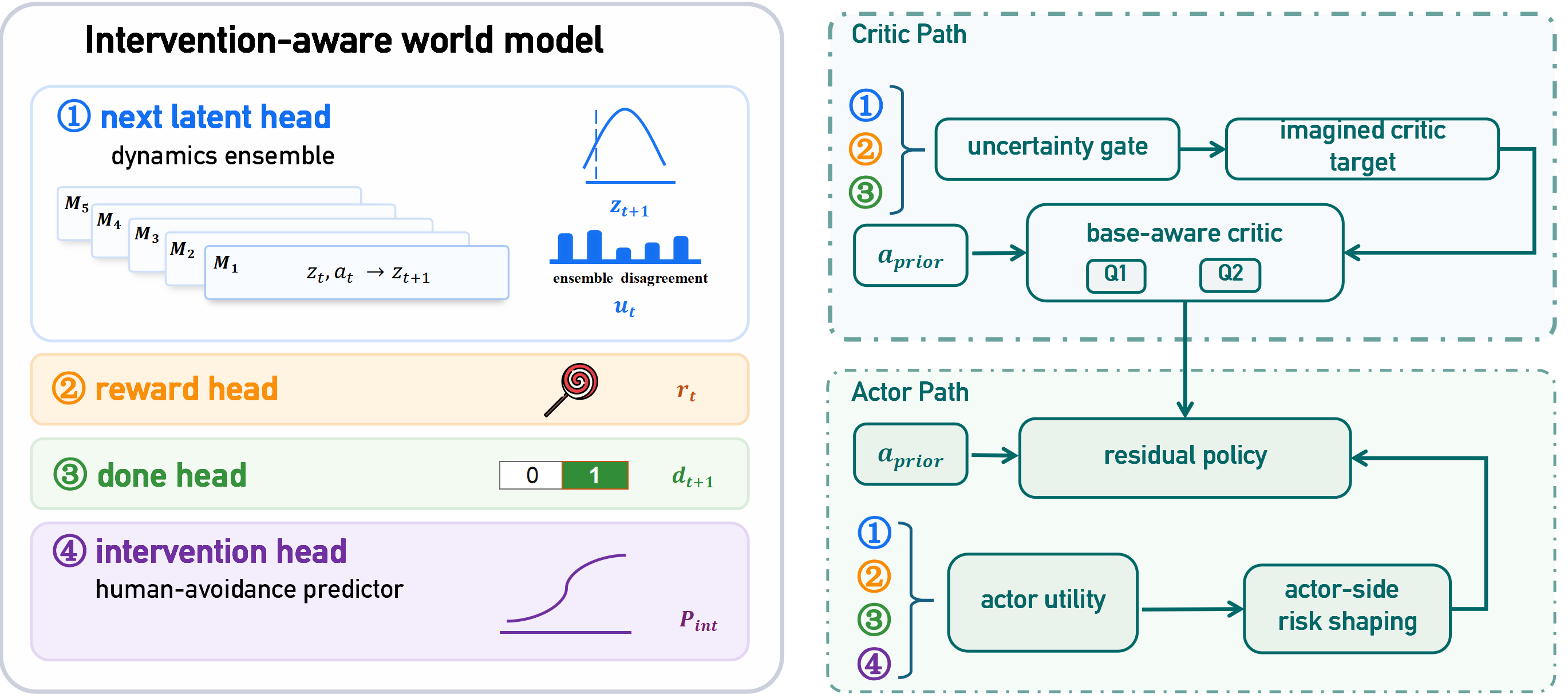}
{\small\caption{\textbf{Intervention-aware world model.} Dynamics, reward, and termination heads support uncertainty-gated critic updates, while the intervention head provides actor-side risk shaping.}\label{fig:wm-diagram}}
\end{figure}

The world model operates on the critic encoder's detached latent $z_t = E(s_t)$. Given $(z_t,a_t)$, a dynamics ensemble predicts $\hat{z}_{t+1}^{(m)} = z_t + g_m(z_t,a_t)$ for $m\!=\!1,\dots,M$, while scalar heads predict reward probability $\hat{r}_{t+1}$, termination $\hat{d}_{t+1}$, and next-step intervention probability $\hat{p}^\mathrm{intv}_{t+1}$. Ensemble disagreement $u(z_t,a_t)\!=\!\mathrm{std}_m\hat{z}_{t+1}^{(m)}$ provides an epistemic uncertainty estimate. The heads are trained jointly on replay with
\begin{equation}
\mathcal{L}_\mathrm{WM} = \mathrm{MSE}(\hat{z}_{t+1},z_{t+1}) + \mathrm{BCE}(\hat{r}_{t+1},\mathbf{1}[r_{t+1}>0]) + \mathrm{BCE}(\hat{d}_{t+1},d_{t+1}) + m_t\!\cdot\!\mathrm{BCE}(\hat{p}^\mathrm{intv}_{t+1},i_{t+1})
\end{equation}
where $m_t\!\in\!\{0,1\}$ is a label-availability mask for the next-step pedal signal (set to $0$ for replay segments without logged takeover state), and $\mathbf{1}[r_{t+1}>0]$ is the binary target for the reward head. The intervention target $i_{t+1}$ is the takeover label at the predicted next step, so the head is trained as a forward predictor rather than a per-step classifier of the current state.

\textbf{Critic path.} The critic is \emph{base-aware}: it consumes the executed action together with the behavior-prior action as concatenated features, $Q(z, [\mathbf{a}, \mathbf{a}^{\mathrm{BC}}])$, so each value estimate is conditioned on the prior action being corrected. For each real transition, the WM produces $(\hat{z}_{t+1}, \hat{r}, \hat{d})$ and an uncertainty gate $g_u = \sigma((\tau - u)/\beta)$. Using the actor proposal $\tilde{\mathbf{a}}_t\sim\pi_\theta(\mathbf{o}_t)$ and the current prior action $\mathbf{a}^{\mathrm{BC}}_t$, the imagined target and critic auxiliary loss are
\begin{equation}
\begin{aligned}
y^\mathrm{imag}
&= \hat{r} + \gamma(1-\hat{d})\min_k Q_k^\mathrm{tgt}(\hat{z}_{t+1},[\tilde{\mathbf{a}}_t, \mathbf{a}^{\mathrm{BC}}_t]),\\
\mathcal{L}^{\mathrm{WM}}_{\mathrm{critic}}
&= \mathbb{E}_{\mathcal{D}}\!\left[
g_u\,\bigl\|Q_\phi(z_t,[\tilde{\mathbf{a}}_t,\mathbf{a}^{\mathrm{BC}}_t])-y^\mathrm{imag}\bigr\|_2^2
\right].
\end{aligned}
\end{equation}
The full critic objective adds this term to the standard real-transition Bellman loss: $\mathcal{L}_{\mathrm{critic}}=\mathcal{L}^{\mathrm{TD}}_{\mathrm{critic}}+\lambda_c\mathcal{L}^{\mathrm{WM}}_{\mathrm{critic}}$. The takeover-probability head is intentionally excluded from this Bellman target, so operator habits do not directly rewrite the critic's return estimate.

\textbf{Actor path.} The actor consumes the intervention head through a model-derived utility:
\begin{equation}
J_\mathrm{WM}(z,a) = \hat{r} - \alpha_d\hat{d} - \alpha_i\hat{p}^\mathrm{intv} - \alpha_u u(z,a),
\label{eq:actor_wm_utility}
\end{equation}
The corresponding auxiliary actor loss is
\begin{equation}
\mathcal{L}^{\mathrm{WM}}_{\mathrm{actor}}
= -\mathbb{E}_{\mathcal{D}}\!\left[g_u\,J_\mathrm{WM}(z_t,\pi_\theta(o_t))\right],
\label{eq:actor_wm_loss}
\end{equation}
The total actor objective is $\mathcal{L}^{\mathrm{total}}_{\mathrm{actor}}=\mathcal{L}_{\mathrm{actor}}+\lambda_a\mathcal{L}^{\mathrm{WM}}_{\mathrm{actor}}$, with $\mathcal{L}_{\mathrm{actor}}$ from Eq.~\ref{eq:actor_loss}. Model gradients are detached, so this loss shapes the actor without distorting WM training. The $\hat{r}$ term uses the calibrated reward-head output, while the termination, intervention, and uncertainty terms use raw probabilities or disagreement. Setting $\alpha_i=0$ cleanly removes the only HIL-specific actor penalty.

\textbf{Risk-shaping intuition.} Eq.~\ref{eq:actor_wm_utility} gives the actor three penalties: termination for drop/slip failures, intervention probability for states the operator would take over, and ensemble disagreement for epistemic outliers. The new term is $\hat{p}^\mathrm{intv}$: because operators usually intervene before terminal failure, it supplies an earlier actor-side risk signal than $\hat{d}$ alone. WM gradients are detached in both actor and critic use, so shaping errors affect exploration without training the WM through its own predictions.

\textbf{Design choice: actor shaping over reward bonus.} We use $\hat{p}^\mathrm{intv}$ as actor-side shaping rather than an environment reward bonus because it predicts operator behavior, not task cost. Putting it into the reward would propagate operator habits through Bellman backups and entangle WM bias with value learning.

\section{Experiments}
\label{sec:experiments}

We organize evaluation around three questions: \textbf{Q1:} final task success; \textbf{Q2:} operator-controlled training steps; and \textbf{Q3:} whether the gain comes from actor-side use of the intervention predictor.

\subsection{Tasks}
\label{sec:exp-setup}

\begin{figure}[h!]
\centering
\includegraphics[width=\linewidth]{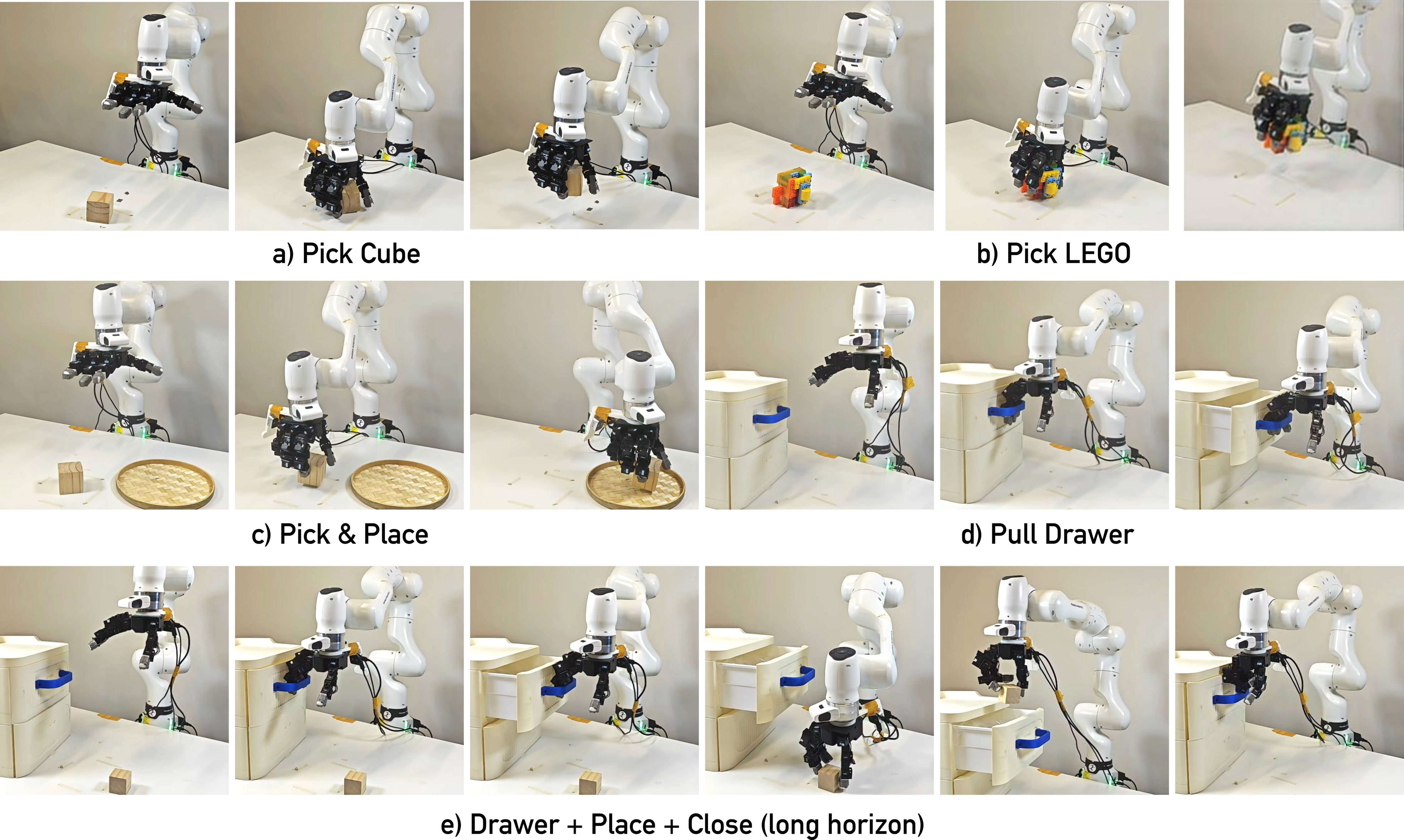}
{\small\caption{\textbf{Real-robot task suite.} Five LEAP-Hand tasks spanning grasping, placement, prismatic manipulation, and the three-stage Long Horizon sequence.}\label{fig:task_strip}}
\end{figure}

Our hardware setup comprises a Franka FR3 arm with a 16-DoF LEAP Hand~\cite{shaw2023leaphand} as the end-effector, two RGB cameras (one wrist-mounted, one third-person), and the tri-channel teleoperation rig described in Section~\ref{sec:method-teleop}. We evaluate on five tasks of increasing difficulty in the real world, as shown in Fig.~\ref{fig:task_strip}: \textbf{Pick Cube} (convex cube with regular grasp affordances; geometric complement to Pick LEGO), \textbf{Pick LEGO} (irregular non-convex object with narrow grasp affordances), \textbf{Pick \& Place} (grasp and placement at a target region), \textbf{Pull Drawer} (handle grasp and prismatic pull), and \textbf{Pull Drawer + Place + Close} (\textbf{Long Horizon}; pull the drawer open, place an object inside, close it).

\subsection{Baselines}
\label{sec:exp-baselines}
We frame the comparison as a \emph{progressive ladder}: four policies share the same observation and action space, and each row adds one mechanism to the previous one:
\begin{equation*}
\text{Behavior prior}
\;\rightarrow\;
\text{ResFiT}
\;\rightarrow\;
\text{Residual RL (model-free)}
\;\rightarrow\;
\mathbf{WHIRL}.
\end{equation*}
\textbf{Behavior prior}~\cite{zhao2023act} $\pi_\mathrm{BC}$: frozen chunk-based imitation policy on $\mathcal{D}_\mathrm{demo}$. \textbf{ResFiT}~\cite{ankile2025resfit}: autonomous residual RL on top under a sparse binary task reward, no operator interventions or world model. \textbf{Residual RL (model-free)}: our HIL baseline, which adds takeover-aware critic training to the residual actor but uses no world model; it follows the core residual-HIL mechanism used by HIL-SERL~\cite{luo2024hilserl} and SiLRI~\cite{silri2025}. \textbf{WHIRL (ours)}: the full pipeline.

\subsection{Metrics}
\label{sec:exp-metrics}
We report one headline trace metric and one offline success-rate measurement, and define the raw per-episode signal from which the trace is computed.
\textbf{Rolling intervention fraction (5-episode, step-weighted)} is the headline trace metric. For episode $i$ with $u_i$ operator-controlled steps and length $T_i$, $\rho_k = \bigl(\sum_{i=k-4}^{k} u_i\bigr) / \bigl(\sum_{i=k-4}^{k} T_i\bigr)$ measures the recent fraction of robot steps under human control; Figure~\ref{fig:main} plots this smoothed trace.
\textbf{Per-episode intervention fraction} $u_i/T_i$ is the raw signal used to compute $\rho_k$.
\textbf{Real-robot success rate}: per-method, per-task fraction over $N$ rollouts of the converged policy under the same in-region randomization as training, reported in Table~\ref{tab:main}.
Only the two HIL residual methods face an active takeover channel; BC and ResFiT run with the pedal disabled, so Figure~\ref{fig:main} contrasts Residual~RL (WHIRL w/o WM) and WHIRL.

\subsection{Training and evaluation details}
\label{sec:training-eval}
Each method starts from the same action-chunk behavior-cloning prior~\cite{zhao2023act} fit to {15--30 teleoperated demonstrations per task} and frozen during online learning, then trains online until trace metrics plateau on each (task, method) cell. We evaluate \emph{in-region generalization}: object-pose, approach-offset, and task-specific handle-position variation within the same workspace markers used for demonstrations. During training, one operator follows a fixed rule shared across methods: take over only when the policy is about to violate workspace bounds or drop the object, and release once the failure mode is averted. For Long Horizon, evaluation is end-to-end rather than stage-wise: the policy must open the drawer, place the object, and close the drawer in a single rollout without per-stage resets.

\textbf{Fully autonomous success-rate evaluation.} Table~\ref{tab:main} is measured on held-out deployments with the pedal disabled and the operator present only for emergency stop, using the checkpoint closest to the per-task nominal budget for each cell. Timeouts, drops, and workspace violations count as failures. This deployment is stricter than the HIL training trace because the operator may emergency-stop but cannot rescue or correct the trajectory.

\textbf{Evaluation rigor under a single seed.} Real-robot training cost makes seed averaging prohibitive, so each (task, method) cell uses one seed and one operator, as in prior real-robot RL studies~\cite{luo2024serl, luo2024hilserl, silri2025}. We compensate with randomized in-region resets, step-weighted rolling traces, and held-out autonomous success evaluation. The four-method ladder shares the same seed/operator pair within each task, so row-to-row differences isolate the added mechanism.

\subsection{Main Results}

\noindent
\begin{minipage}[t]{0.46\linewidth}
\vspace{0pt}
\noindent\textbf{Q1.~Final task success.}
WHIRL improves over the strongest baseline (Residual~RL model-free) on every task by $+15$--$30$ percentage points (Table~\ref{tab:main}). Fisher's exact reaches $p\!<\!0.05$ on the three pick tasks; Pull Drawer and Long Horizon are suggestive at $p\!\approx\!0.08$. ResFiT and Residual~RL differ by at most one trial per task, so the consistent gap appears when the world-model row is added.
\end{minipage}\hfill
\begin{minipage}[t]{0.51\linewidth}
\vspace{0pt}
\centering
\includegraphics[width=0.9\linewidth]{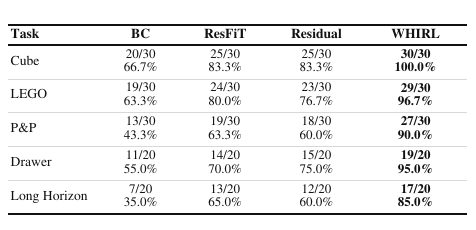}
\vspace{-0.9em}
{\small\captionof{table}{\textbf{Performance.} Each cell shows successes/rollouts over success rate.}\label{tab:main}}
\end{minipage}
\par\vspace{0.3em}

\noindent
\begin{minipage}{\linewidth}
\centering
\includegraphics[width=\linewidth]{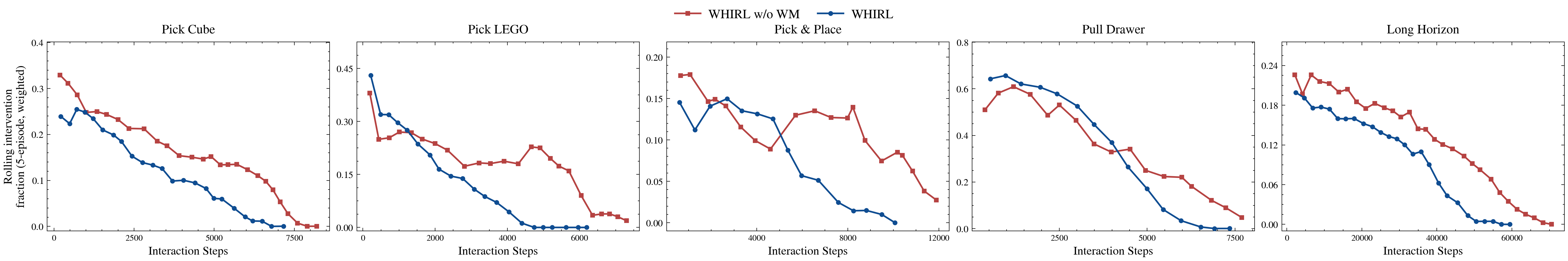}
\vspace{-0.6em}
{\small\captionof{figure}{\textbf{Main result: rolling intervention fraction (lower is better).} Step-weighted rolling-5 intervention fraction $\rho_k$ across five real-robot tasks. WHIRL stays below the no-WM baseline and reaches the near-zero regime earlier; success rates are in Table~\ref{tab:main}.}\label{fig:main}}
\end{minipage}
\vspace{-0.5em}

\paragraph{Q2: Operator intervention burden.}
WHIRL stays below the model-free baseline on the rolling-5 trace for every task (Figure~\ref{fig:main}). On pick tasks, both methods eventually reach $\rho_k\!=\!0$, but WHIRL gets there earlier. On Pull Drawer and Long Horizon, the step-weighted curve separates the methods despite noisy raw traces. The step-weighted view matters because a brief pedal tap and a long recovery should not count equally. The largest relative drop is on Pull Drawer, where WHIRL reduces the final rolling fraction from the baseline's $0.113$ to $0.018$ ($84\%$). Since autonomous success improves in the same direction, lower intervention burden is not caused by early termination or giving up. Operationally, the operator shifts from frequent recovery to rare boundary correction. Under the fixed operator protocol, a lower step-weighted fraction means fewer prolonged rescue segments, not merely fewer pedal events. On Pull Drawer, the remaining interventions are brief reach-margin corrections rather than sustained recoveries.

\noindent
\begin{minipage}[t]{0.42\linewidth}
\vspace{0pt}
\centering
\includegraphics[width=0.80\linewidth]{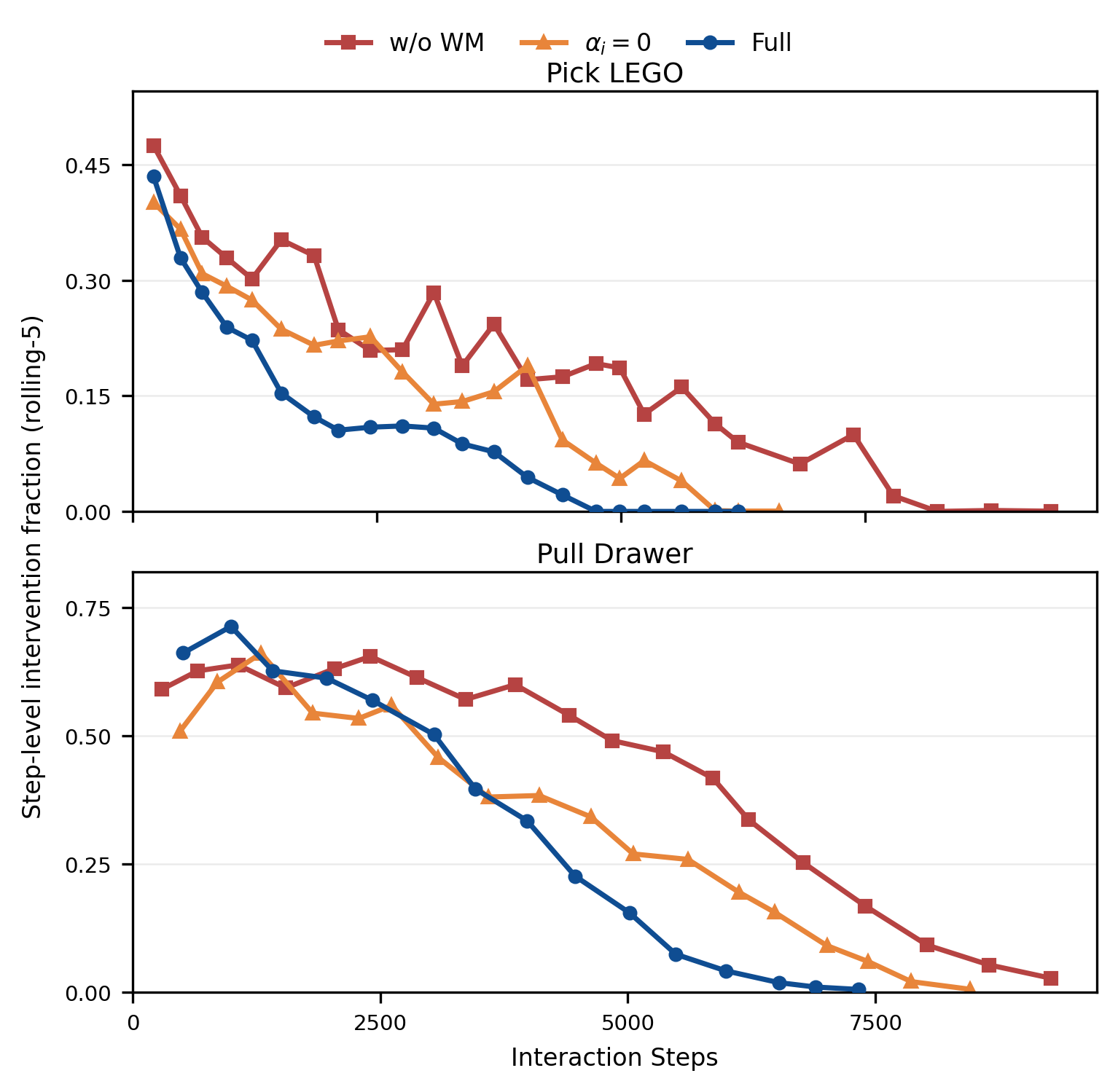}
\vspace{-0.5em}
{\small\captionof{figure}{\textbf{Mechanism ablation.} Removing actor-side intervention shaping or the WM delays convergence.}\label{fig:ablation}}
\end{minipage}\hfill
\begin{minipage}[t]{0.48\linewidth}
\vspace{0pt}
\paragraph{Q3: Source of the gain.}\label{sec:exp-ablation}
We isolate the mechanism on Pick LEGO (multi-finger grasping) and Pull Drawer (prismatic contact). \emph{WHIRL ($\alpha_i\!=\!0$)} trains the world model but blocks actor use of the intervention head; \emph{WHIRL w/o WM} removes the model pathway. Full WHIRL reaches the near-zero intervention regime first on both tasks. The $\alpha_i\!=\!0$ curve stays above full WHIRL even though the critic-facing scaffold remains, so the gain is not just auxiliary BCE training. WHIRL w/o WM is slowest, confirming that the model pathway also carries part of the burden. Pull Drawer shows a brief early cost before crossover, consistent with risk shaping helping only after the intervention head fits enough replay.
\end{minipage}
\par\vspace{0.5em}

\section{Limitations}
\label{sec:discussion}

WHIRL remains bounded by real-robot HIL constraints. Evaluation covers \emph{in-region} pose and approach randomization rather than new objects, layouts, or fine in-hand manipulation. Training uses one seed and one operator per cell. The Pull Drawer and Long Horizon success gaps remain suggestive at current sample sizes, with the mechanism ablation providing complementary evidence. The intervention head also inherits the takeover threshold and habits of the labeling operator, so deployment with a substantially more or less cautious operator may require recalibration. Natural next settings include precision in-hand reorientation and fingertip gaiting on different hand morphologies or tactile-equipped platforms, where failures may arise from subtle contact-quality changes rather than gross drops.

\section{Conclusion}
\label{sec:conclusion}

WHIRL turns takeovers into reusable risk predictions for actor-side shaping. Across five real-robot dexterous tasks, it improves autonomous success by 15--30 percentage points over the strongest baseline (Residual~RL model-free) and reduces operator-controlled training steps by up to 84\%. The result suggests a practical recipe for dexterous HIL-RL: keep corrections on the real robot, learn when the operator would intervene, and route that prediction to the actor rather than the Bellman target. More broadly, WHIRL treats operator attention as reusable supervision instead of a one-off rescue signal. \textbf{To our knowledge, WHIRL is the first HIL-RL system to use per-state takeover prediction as an actor-side risk signal on a 16-DoF dexterous hand.}

\clearpage
\bibliography{example}

\appendix

\section{Hand Retargeter Implementation Details}
\label{app:retargeter}

This appendix gives the closed-form expressions and solver hyperparameters of the hybrid hand retargeter summarized in Section~\ref{sec:method-teleop}. Let $\mathbf{g}_t$ denote the calibrated Manus glove state and $\mathbf{q}^{L}_t \in \mathbb{R}^{16}$ the commanded LEAP joints; superscripts $L$ and $G$ refer to the LEAP and glove frames, respectively.

\paragraph{Non-thumb fingers (per-joint affine map).}
For each finger $f \in \mathcal{F}_{\mathrm{map}} = \{\mathrm{index}, \mathrm{middle}, \mathrm{ring}\}$ and joint $j$, the commanded LEAP joint is
\begin{equation}
    q^{L}_{f,j,t} =
    \mathrm{clip}\!\left(\alpha_{f,j}\, g_{f,j,t} + \beta_{f,j},\;
    q^{L}_{f,j,\min},\; q^{L}_{f,j,\max}\right),
    \label{eq:finger_mapping}
\end{equation}
with scalar gain $\alpha_{f,j}$ and offset $\beta_{f,j}$ fit once per operator from three reference postures (fully open, half-closed, fully closed) by two-parameter, three-point least squares; the closed-form solution is recomputed offline rather than online.

\paragraph{Thumb (workspace fingertip IK with opposition alignment).}
For the 4-DoF thumb chain, we first define three residuals:
\begin{equation}
\begin{aligned}
    \mathbf{r}_x(\mathbf{q})
    &= \mathbf{x}^{L}_{\mathrm{th}}(\mathbf{q})
       - \mathbf{T}_{GL}\mathbf{x}^{G}_{\mathrm{th}}(\mathbf{g}_t),\\
    \mathbf{r}_o(\mathbf{q})
    &= \hat{\mathbf{n}}^{L}(\mathbf{q})
       - \mathbf{R}_{GL}\hat{\mathbf{n}}^{G}(\mathbf{g}_t),\\
    \mathbf{r}_s(\mathbf{q})
    &= \mathbf{q} - \mathbf{q}^{L}_{\mathrm{th},t-1}.
\end{aligned}
\label{eq:thumb_residuals}
\end{equation}
The commanded thumb posture is then the constrained minimizer
\begin{equation}
    \mathbf{q}^{L}_{\mathrm{th},t}
    =
    \arg\min_{\mathbf{q}\in[\mathbf{q}_{\min},\mathbf{q}_{\max}]}
    \|\mathbf{r}_x(\mathbf{q})\|_2^2
    + \lambda_o\|\mathbf{r}_o(\mathbf{q})\|_2^2
    + \lambda_{\mathrm{smooth}}\|\mathbf{r}_s(\mathbf{q})\|_2^2,
    \label{eq:thumb_ik}
\end{equation}
where $\mathbf{r}_x$ tracks the thumb fingertip target, $\mathbf{r}_o$ aligns the thumb--index opposition direction, and $\mathbf{r}_s$ smooths frame-to-frame commands. Here $\mathbf{x}^{L}_{\mathrm{th}}$ is the LEAP thumb forward kinematics, $\mathbf{x}^{G}_{\mathrm{th}}$ the glove-derived fingertip target, $(\mathbf{T}_{GL}, \mathbf{R}_{GL})$ the translational and rotational parts of the one-shot palm-to-palm calibration from $G$ to $L$, and $\hat{\mathbf{n}}^{L}(\mathbf{q}) = (\mathbf{x}^{L}_{\mathrm{idx}} - \mathbf{x}^{L}_{\mathrm{th}}(\mathbf{q}))/\|\cdot\|$, $\hat{\mathbf{n}}^{G}(\mathbf{g}_t) = (\mathbf{x}^{G}_{\mathrm{idx}} - \mathbf{x}^{G}_{\mathrm{th}})/\|\cdot\|$ are the thumb-tip\,$\to$\,index-tip unit vectors in the two frames. The LEAP index-tip $\mathbf{x}^{L}_{\mathrm{idx}}$ is taken from the affine command of Eq.~\ref{eq:finger_mapping} at the current frame and held fixed during the thumb solve, so $\hat{\mathbf{n}}^{L}$ depends on $\mathbf{q}$ only through the thumb-tip term.

\textit{Gauss--Newton derivation.}
Let $J = \partial \mathbf{x}^{L}_{\mathrm{th}}/\partial \mathbf{q} \in \mathbb{R}^{3\times 4}$ be the position Jacobian of the LEAP thumb tip, $\mathbf{e} = \mathbf{T}_{GL}\mathbf{x}^{G}_{\mathrm{th}}(\mathbf{g}_t) - \mathbf{x}^{L}_{\mathrm{th}}(\mathbf{q}_k)$, $\mathbf{m} = \mathbf{R}_{GL}\hat{\mathbf{n}}^{G}(\mathbf{g}_t)$, $\mathbf{v} = \mathbf{x}^{L}_{\mathrm{idx}} - \mathbf{x}^{L}_{\mathrm{th}}(\mathbf{q}_k)$, and let $\mathbf{P} = I_3 - \hat{\mathbf{n}}^{L}\hat{\mathbf{n}}^{L\top}$ be the projector orthogonal to $\hat{\mathbf{n}}^{L}$. The opposition residual is $\mathbf{r}_o = \hat{\mathbf{n}}^{L}(\mathbf{q}) - \mathbf{m} \in \mathbb{R}^3$ with Jacobian $\partial \mathbf{r}_o/\partial \mathbf{q} = -(1/\|\mathbf{v}\|)\,\mathbf{P}\,J$. Linearizing all three terms of Eq.~\ref{eq:thumb_ik} around $\mathbf{q}_k$ and using $\mathbf{P}\hat{\mathbf{n}}^{L} = 0$ to simplify the gradient, the damped Gauss--Newton step solves the normal equations
\begin{equation}
    \Bigl(J^{\!\top}\!J \;+\; \frac{\lambda_o}{\|\mathbf{v}\|^2} J^{\!\top}\mathbf{P}\,J \;+\; (\lambda_{\mathrm{smooth}} + \mu^{2}) I_4\Bigr)\,\Delta \mathbf{q}_k
    \;=\;
    J^{\!\top}\mathbf{e} \;-\; \frac{\lambda_o}{\|\mathbf{v}\|}\,J^{\!\top}\mathbf{P}\,\mathbf{m} \;+\; \lambda_{\mathrm{smooth}}\bigl(\mathbf{q}^{L}_{\mathrm{th},t-1} - \mathbf{q}_k\bigr),
    \label{eq:thumb_gn}
\end{equation}
with Levenberg--Marquardt damping $\mu = 2\!\times\!10^{-2}$ for stability near singularities of the 4-DoF chain. The negative sign on the opposition RHS term follows from $\partial \hat{\mathbf{n}}^{L}/\partial \mathbf{q} = -(1/\|\mathbf{v}\|)\,\mathbf{P}\,J$ and $\mathbf{P}\hat{\mathbf{n}}^{L}=0$: collecting the gradient of $\lambda_o\|\hat{\mathbf{n}}^{L}-\mathbf{m}\|^2$ at $\mathbf{q}_k$ and applying the GN step $\Delta\mathbf{q} = -H^{-1}\nabla L$ yields $-(\lambda_o/\|\mathbf{v}\|) J^\top \mathbf{P}\,\mathbf{m}$. The opposition contribution to the Hessian is rank-$\le 2$ (it lives in the 2-D subspace orthogonal to $\hat{\mathbf{n}}^{L}$), so it never inflates the solve in directions along the opposition axis.

\paragraph{Solver schedule.}
Up to 40 iterations per control step, step size $0.3$, per-joint clipping to LEAP limits, temporal smoothing weight $\lambda_{\mathrm{smooth}} = 5\!\times\!10^{-2}$, opposition weight $\lambda_o = 2\!\times\!10^{-1}$. Solver state is warm-started from $\mathbf{q}^{L}_{\mathrm{th},t-1}$, and we early-stop at $\|\mathbf{e}\|_2 < 10^{-3}$\,m; under typical glove rates the solver converges in fewer than 10 iterations.

\paragraph{Intervention rebase formula.}
Let $R(\cdot)$ denote the per-frame retargeter (Eq.~\ref{eq:finger_mapping} on the non-thumb fingers, Eq.~\ref{eq:thumb_ik} on the thumb), evaluated for this rebase as a memoryless function of the current glove sample so that the rebase anchor and delta are deterministic in $\mathbf{g}_t$; the running solver of Eq.~\ref{eq:thumb_ik} still retains its temporal warm-start during continuous control. After a takeover at time $t_0$ the command is
\begin{equation}
    \mathbf{q}^{L,\mathrm{cmd}}_t
    = \mathbf{q}^{L,\mathrm{robot}}_{t_0} \;+\; \bigl(R(\mathbf{g}_t) - R(\mathbf{g}_{t_0})\bigr),
    \label{eq:teleop_rebase}
\end{equation}
so the first commanded pose equals the current robot hand pose and subsequent commands carry only the joint-space \emph{increment} of the retargeted posture since $t_0$.

\section{World Model and RL Hyperparameters}
\label{app:hyperparams}

Table~\ref{tab:wm-hp} lists the world-model and critic/actor hyperparameters used in all real-robot runs; values are shared across the five tasks unless noted. Ablations keep these values except for their named removals (e.g.\ $\alpha_i\!=\!0$ or no WM). The daggered residual scales are paired with explicit L2 penalties on arm and hand residuals; these penalties keep early online updates local around the frozen behavior prior and are annealed once the residual policy reaches stable task execution. The exact per-task residual-penalty schedule is fixed before training and included in the task configuration files of the supplementary code.

\paragraph{Real-transition critic loss.}
The backbone critic is trained on real replay transitions with the standard clipped-double-Q Bellman target
\begin{equation}
\begin{aligned}
y^{\mathrm{TD}}_t
&= r_{t+1} + \gamma(1-d_{t+1})
\min_k Q_k^{\mathrm{tgt}}\!\left(z_{t+1},[\tilde{\mathbf{a}}_{t+1},\mathbf{a}^{\mathrm{BC}}_{t+1}]\right),\\
\mathcal{L}^{\mathrm{TD}}_{\mathrm{critic}}
&= \mathbb{E}_{\mathcal{D}}\!\left[
\bigl\|Q_\phi(z_t,[\mathbf{a}_t,\mathbf{a}^{\mathrm{BC}}_t])-y^{\mathrm{TD}}_t\bigr\|_2^2
\right],
\end{aligned}
\label{eq:td_critic_appendix}
\end{equation}
where $\tilde{\mathbf{a}}_{t+1}\!\sim\!\pi_\theta(\cdot\,|\,z_{t+1})$ is a reparameterized sample from the current actor, $\mathbf{a}^{\mathrm{BC}}_{t+1}$ is the frozen behavior-prior action at the next state, and $z_{t+1}$ is the encoded real next observation. The standard SAC entropy term $-\alpha\log\pi_\theta(\tilde{\mathbf{a}}_{t+1}\,|\,z_{t+1})$ is absorbed into $y^{\mathrm{TD}}_t$ as in clipped-double-Q SAC and is omitted from Eq.~\ref{eq:td_critic_appendix} for brevity. Section~\ref{sec:method-wm} adds the uncertainty-gated world-model auxiliary target on top of this loss.

\begin{table}[h]
\centering
\small
\setlength{\tabcolsep}{6pt}
\begin{tabular}{l l l}
\toprule
Component & Hyperparameter & Value \\
\midrule
\multirow{6}{*}{World model}
  & Latent dimension $d_z$                                & 256 \\
  & Head MLP width / depth                                & $2\times512$ hidden \\
  & Dynamics ensemble size $M$                            & 5 \\
  & Optimizer / learning rate                             & Adam / $1\!\times\!10^{-4}$ \\
  & Batch size                                            & 128 \\
  & Calibrated reward $(R_\mathrm{succ}, R_\mathrm{step})$ & $(10.0, -0.01)$ \\
\midrule
\multirow{5}{*}{Actor \& Critic}
  & Discount $\gamma$                                     & 0.97 \\
  & Imagined critic weight $\lambda_c$                    & 0.05 \\
  & Actor WM weight $\lambda_a$                           & 0.40 \\
  & Uncertainty soft gate $(\tau, \beta)$                 & $(0.03, 0.01)$ \\
  & Actor utility coeffs.\ $(\alpha_d, \alpha_i, \alpha_u)$     & $(1.0, 0.4, 1.0)$ \\
\midrule
\multirow{3}{*}{Residual RL backbone}
  & Synergy dimension $K$                                 & 3 \\
  & Residual scales (arm / synergy / hand)$^\dagger$      & $(1.5\!\times\!10^{-3},\ 0.25,\ 0.15)$ \\
  & Replay capacity (online / offline / intervention)     & $(10\text{k},\ 500,\ 5\text{k})$ \\
\bottomrule
\end{tabular}
{\small\caption{World-model and residual-RL hyperparameters used for reported WHIRL runs.}\label{tab:wm-hp}}
\end{table}

\section{Design Choices for the Intervention-Aware World Model}
\label{app:design-choice}

\paragraph{1-step deterministic-latent design vs.\ RSSM / Hi-WM.} Each WM head in Section~\ref{sec:method-wm} is a small MLP on the same encoded latent $z_t$: the dynamics ensemble supplies an epistemic mask, the reward and termination heads form a 1-step Bellman target, and the intervention head turns the binary takeover label into forward supervision the actor can consume directly. RSSM-style world models~\cite{hafner2025dreamerv3} learn a recurrent latent and train on imagined rollouts of length 15--16, but on our small-data real-robot budget we found multi-step imagination compounds per-step dynamics error on a 16-DoF hand without sharpening the 1-step Bellman target. Hi-WM~\cite{li2026hiwm} also pairs a world model with HIL, but as a sandbox in which the operator intervenes \emph{inside} the model; WHIRL keeps every correction on the real robot and uses the model only as a 1-step data multiplier (critic) and a per-state risk predictor (actor).

\medskip

WHIRL deliberately routes the takeover-probability head $\hat{p}^\mathrm{intv}$ into the \emph{actor objective alone}, rather than into the environment reward or the critic's Bellman target. This decomposition is the central design choice of the intervention-aware world model and rests on four properties, which we contrast against the natural alternative of using $\hat{p}^\mathrm{intv}$ as a reward bonus $r'(s,a) = r(s,a) - \beta\,\hat{p}^\mathrm{intv}(s,a)$.

\paragraph{Optimality invariance.} $\hat{p}^\mathrm{intv}$ is a learned \emph{predictor} of operator behavior, not a true cost: it reflects \emph{which states an operator chose to take over in the past}, which conflates real failure risk with operator habit, attention budget, and task framing. Used as a reward bonus, this signal changes the optimal policy and can push the actor into over-conservative regimes that avoid even free-space approach states that an operator never actually flagged at deployment time. Used in the auxiliary actor loss of Eq.~\ref{eq:actor_wm_loss}, the same signal biases action updates through $J_\mathrm{WM}$ without redefining the task reward, and $\alpha_i\!\to\!0$ exactly removes intervention shaping.

\paragraph{Decoupling from credit assignment.} A reward bonus propagates through every Bellman backup, so $\hat{p}^\mathrm{intv}$ silently re-weights the entire downstream value function and entangles WM bias with credit assignment. In our decomposition, the critic's Bellman target $y^\mathrm{imag}$ (Section~\ref{sec:method-wm}) consumes only the dynamics, reward, and termination heads; the takeover-probability head touches the actor objective alone. Errors in $\hat{p}^\mathrm{intv}$ therefore cannot leak into $Q$, and the critic's uncertainty-gated targets remain a faithful estimate of return under the calibrated reward.

\paragraph{No conflict with reward calibration.} WHIRL already calibrates its sparse-binary task reward through the reward head, $\hat{r} = \hat{p}_r R_\mathrm{succ} + (1-\hat{p}_r) R_\mathrm{step}$ (Section~\ref{sec:method-wm}, Appendix~\ref{app:hyperparams}). Adding a $-\beta\hat{p}^\mathrm{intv}$ bonus would compete with this calibration on the same scale: $\beta$ and $(R_\mathrm{succ}, R_\mathrm{step})$ would have to be co-tuned per task to prevent the shaping term from dominating success signal. Keeping the takeover term out of $\hat{r}$ preserves a single, calibrated reward channel. The reported $\alpha_i\!=\!0.4$ is therefore a moderate actor-utility weight rather than a redefinition of the environment reward; the mechanism ablation tests its role by setting the weight to zero.

\paragraph{Clean ablation surface.} The two roles of the WM are factored into two orthogonal knobs: removing the world model disables both the uncertainty-gated imagined critic target and actor-side shaping, while setting $\alpha_i\!=\!0$ disables only actor-side intervention shaping and leaves the dynamics/reward/termination scaffold intact. This factoring is exactly what the three-line ablation in Section~\ref{sec:exp-ablation} exploits: \emph{WHIRL w/o WM} measures the value of the WM scaffold, and \emph{WHIRL ($\alpha_i\!=\!0$)} isolates the value of actor consumption of the intervention predictor. A reward-bonus design collapses both pathways into the shaping coefficient $\beta$ and prevents this attribution.

\section{Detailed Training and Evaluation Protocol}
\label{app:protocol}

This appendix expands the protocol summarized in Section~\ref{sec:training-eval}.

\paragraph{Demonstrations and replay seeding.} Each task uses 15--30 teleoperated demonstrations collected with the tri-channel interface of Section~\ref{sec:method-teleop}; the same set both fits the frozen action-chunk behavior-cloning prior~\cite{zhao2023act} and seeds the replay buffer used by the world model. No additional offline data are introduced during online training.

\paragraph{Single seed and within-seed smoothing.} Because each (task, method) cell costs hours of real-robot interaction time, we report a single training seed per cell and rely on trailing 5-episode smoothing (Section~\ref{sec:exp-metrics}) to absorb within-seed noise. Trace metrics in Figures~\ref{fig:main} and~\ref{fig:ablation} tally interaction steps at episode termination, so final logged points may extend past the per-task nominal checkpoint ($\approx$8{,}000 steps for the pick and drawer tasks, $\sim$50{,}000 for Long Horizon); the success-rate table (Table~\ref{tab:main}) instead uses the checkpoint closest to that budget on each cell so that all methods are compared at comparable training cost.

\paragraph{Operator standard operating procedure.} A single human operator runs every (task, method) cell. Because the operator necessarily knows the active method during real-robot training, we hold the intervention rule fixed across conditions to keep labels comparable. Specifically the operator presses the pedal in exactly two situations: (i) the policy is about to violate the Cartesian workspace bounds listed in Table~\ref{tab:workspace}, or (ii) the policy has dropped the object or has visibly lost grasp closure with the object still graspable. The operator releases the pedal as soon as the failure mode is averted. No preemptive corrections, posture cleanup, or smoothing demonstrations are injected during online runs.

\paragraph{Implications for variance reporting.} The combination of a single seed and a single operator with method awareness means our trace metrics lack across-seed bands. We rely on three complementary signals to mitigate this: (i) the step-weighted rolling intervention fraction (Figure~\ref{fig:main}, Section~\ref{sec:exp-metrics}) is less volatile than the raw per-episode fraction (Appendix~\ref{app:raw-traces}) and is therefore more discriminative when the operator's role shrinks; (ii) the success-rate evaluation in Table~\ref{tab:main} does not depend on per-step takeover decisions and so removes operator bias from the headline outcome; and (iii) the four-row baseline ladder of Table~\ref{tab:main} lets each cell share the same operator and seed pair, so the \emph{differences} between rows isolate the corresponding mechanism even if the absolute level reflects a particular run.

\paragraph{Statistical analysis for Table~\ref{tab:main}.}\label{app:table1-stats}
We report two complementary statistics per task: a Wilson score 95\% confidence interval for WHIRL's success rate (small-$N$ appropriate; binomial), and a one-sided uncorrected Fisher's exact $p$-value for the WHIRL-vs-Residual\,RL contrast (testing whether WHIRL succeeds strictly more often than the strongest no-WM baseline given the row/column totals). No multiple-comparison correction is applied (five tasks), so the reported $p$-values should be read as per-task evidence rather than as a family-wise significance.
\begin{center}\small
\begin{tabular}{l c c}
\toprule
Task & WHIRL Wilson 95\% CI & Fisher one-sided $p$ (WHIRL vs.\ Residual\,RL) \\
\midrule
Pick Cube     & $[89, 100]\%$ & $0.026$ \\
Pick LEGO     & $[83, 99]\%$ & $0.026$ \\
Pick \& Place & $[74, 97]\%$ & $0.008$ \\
Pull Drawer   & $[76, 99]\%$ & $0.091$ \\
Long Horizon  & $[64, 95]\%$ & $0.078$ \\
\bottomrule
\end{tabular}\end{center}
The three pick tasks (each at $N\!=\!30$) reach $p\!<\!0.05$; Pull Drawer and Long Horizon (each at $N\!=\!20$) are in the suggestive regime ($p\!\approx\!0.08$--$0.09$), consistent with the smaller sample size on those two tasks rather than a smaller underlying effect. The wider lower bound on Long Horizon's WHIRL CI ($64\%$) compared with Pull Drawer ($76\%$) reflects Long Horizon's $\hat{p}\!=\!17/20\!=\!85\%$ being further from $1$ than Pull Drawer's $19/20\!=\!95\%$ at the same $N$.

\paragraph{Baseline implementation rationale.}\label{app:baseline-rationale}
The \emph{Residual RL (model-free)} baseline (Section~\ref{sec:exp-baselines}) is our in-house implementation of the HIL residual-actor + intervention-aware-critic mechanism shared by HIL-SERL~\cite{luo2024hilserl} and SiLRI~\cite{silri2025}, rather than the published codebases. The motivation is implementation-difference isolation: the published HIL-SERL and SiLRI codebases target different end-effectors (parallel-jaw grippers and bimanual setups) and different observation spaces from our 16-DoF LEAP Hand + dual-RGB setup, so running them verbatim would conflate any gap with hardware/observation mismatch rather than algorithmic content. By sharing the behavior prior, residual actor, synergy-space residual parameterization, replay buffer, and HIL pedal channel with WHIRL, we make the ResFiT~$\to$~Residual RL (model-free)~$\to$~WHIRL ladder a clean three-step isolation of (i) autonomous residual RL on top of the prior, (ii) adding the HIL pedal channel into the critic, and (iii) adding the intervention-aware world model. Each row differs from the previous by exactly one mechanism, which is what the row-to-row gaps in Table~\ref{tab:main} report.

\section{Raw Per-Episode Intervention Traces}
\label{app:raw-traces}

Figure~\ref{fig:main} in the main paper reports the step-weighted 5-episode rolling intervention fraction $\rho_k$. Figure~\ref{fig:raw-traces} below shows the underlying per-episode fraction $u_i/T_i$ from which $\rho_k$ is deterministically derived, for the same five tasks and the same WHIRL vs.\ WHIRL w/o WM contrast. The raw signal is noisier episode-to-episode (single-episode fluctuations in $T_i$ amplify into $u_i/T_i$), which is exactly the variance the rolling-5 metric absorbs; both curves converge to the same near-zero regime once the policy stops requiring operator help. We keep this raw view in the appendix because it confirms the rolling-5 trace is not hiding a slow drift, while leaving the main paper figure uncluttered.

\begin{figure}[h!]
\centering
\includegraphics[width=\linewidth]{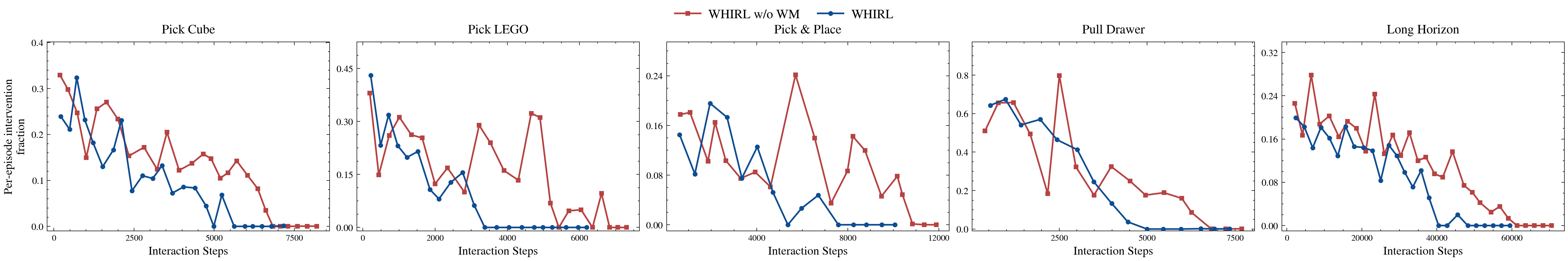}
{\small\caption{\textbf{Raw per-episode intervention fraction $u_i/T_i$ (lower is better).} Same five tasks and same WHIRL (blue, circles) vs.\ WHIRL w/o WM (red, squares; $=$ Residual RL model-free) contrast as Figure~\ref{fig:main}. The step-weighted rolling-5 aggregation of this signal is the headline metric in Figure~\ref{fig:main}.}\label{fig:raw-traces}}
\end{figure}

\section{Hardware, Workspace, and Data Collection}
\label{app:hardware}

\paragraph{Robot and sensors.}
A Franka FR3 arm with a 16-DoF LEAP Hand~\cite{shaw2023leaphand} is observed by wrist-mounted D405 and third-person D435 RGB cameras at 30\,Hz. The arm runs at 20\,Hz in Cartesian impedance mode; the LEAP Hand and Manus glove stream at 120\,Hz.

\paragraph{Workspace and reset.}
Table~\ref{tab:workspace} lists the in-region randomization, reset pose, workspace bounds, and horizon for the five tasks; the dagger marks Pick Cube as the convex-object control and Long Horizon as the three-stage task. Reset orientations are end-effector-down (RPY $\approx (180, {-}2, 89)$\,deg for pick-family tasks, $({-}179, 3, 19)$\,deg for drawer-family tasks). Workspace bounds are in the FR3 \texttt{link8} FK frame from raw teleop episodes (p01--p99 plus a $2$\,cm margin). Each behavior prior uses 15--30 demonstrations; online residual RL runs for $\approx$8\,k steps on pick/drawer tasks and $\sim$50\,k on Long Horizon using one RTX 4090.

\begin{center}
\centering
\vspace{-0.3em}
\scriptsize
\setlength{\tabcolsep}{3pt}
\renewcommand{\arraystretch}{0.92}
\begin{tabular}{l c c c c c}
\toprule
Property & Pick Cube$^\dagger$ & Pick LEGO & Pick \& Place & Pull Drawer & Long Horizon$^\dagger$ \\
\midrule
Obj. rand. & $\pm15$\,cm & $\pm15$\,cm & $\pm15$\,cm & $\pm5$\,cm (handle) & $\pm5$\,cm \\
Reset EEF (m) & $(0.34,0.09,0.46)$ & $(0.34,0.09,0.46)$ & $(0.34,0.09,0.46)$ & $(0.42,0.16,0.55)$ & $(0.42,0.16,0.54)$ \\
WS $x$ (m) & $[0.32,0.62]$ & $[0.31,0.58]$ & $[0.32,0.60]$ & $[0.39,0.51]$ & $[0.35,0.58]$ \\
WS $y$ (m) & $[{-}0.08,0.16]$ & $[{-}0.02,0.12]$ & $[{-}0.03,0.31]$ & $[{-}0.03,0.18]$ & $[{-}0.12,0.26]$ \\
WS $z$ (m) & $[0.19,0.48]$ & $[0.19,0.48]$ & $[0.19,0.48]$ & $[0.38,0.57]$ & $[0.21,0.64]$ \\
Horizon & 300\,/\,15\,s & 300\,/\,15\,s & 600\,/\,30\,s & 450\,/\,22.5\,s & 2100\,/\,105\,s \\
\bottomrule
\end{tabular}
{\footnotesize\captionof{table}{Task randomization, reset pose, workspace bounds, and episode horizon.}\label{tab:workspace}}
\end{center}

\begin{center}
\centering
\vspace{-0.5em}
\includegraphics[width=0.40\linewidth]{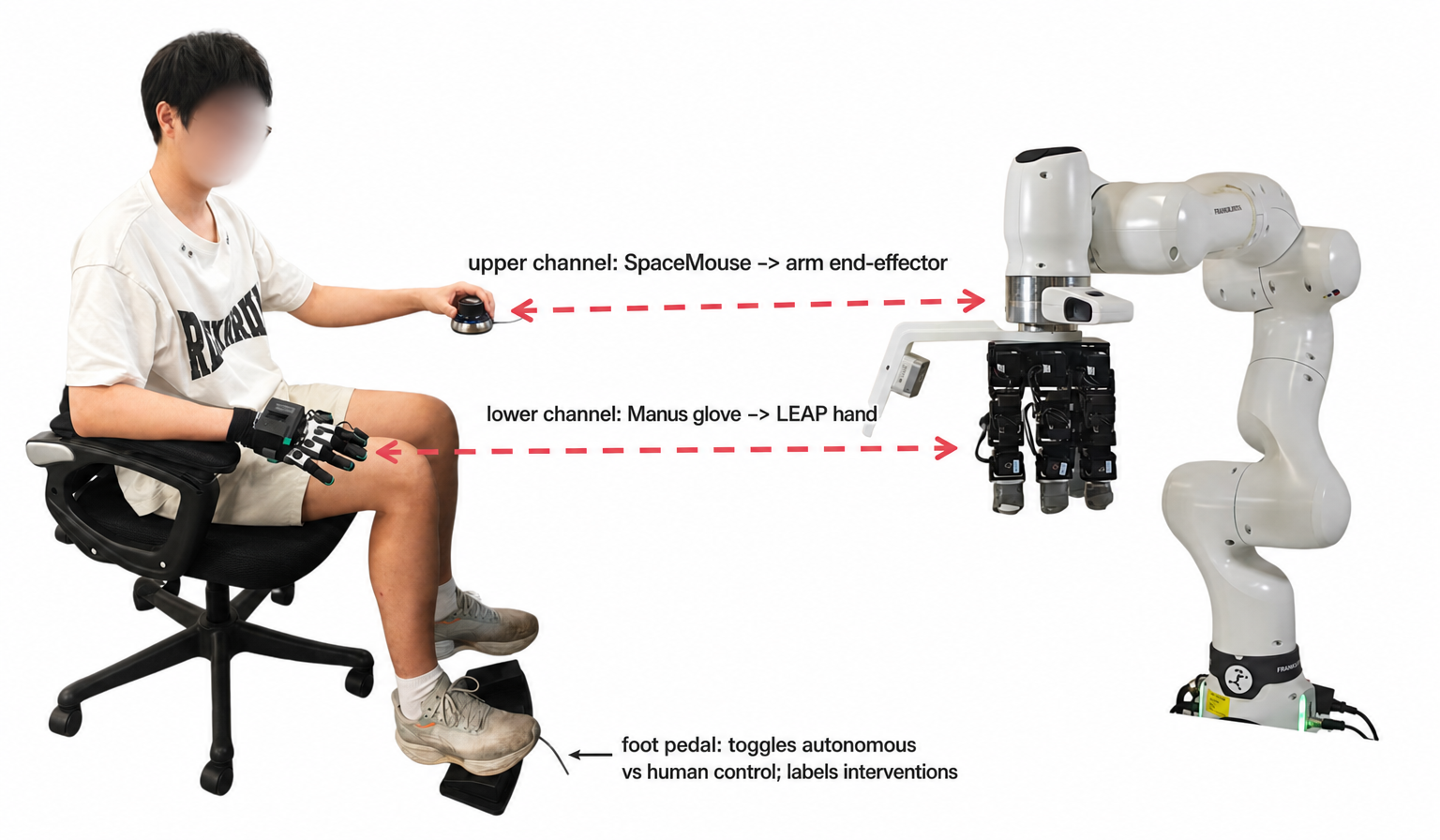}
\vspace{-0.6em}
{\footnotesize\captionof{figure}{\textbf{Tri-channel HIL teleoperation rig:} SpaceMouse (arm), Manus glove (hand), and foot pedal (takeover).}\label{fig:teleoperation}}
\vspace{-0.6em}
\end{center}

\end{document}